%% file: main.tex
\documentclass[11pt]{article}
\usepackage{fontawesome5} 
\usepackage[]{acl}
\usepackage{times}
\usepackage{latexsym}
\usepackage[T1]{fontenc}
\usepackage[utf8]{inputenc}
\usepackage{microtype}
\usepackage{inconsolata}
\usepackage{graphicx}
\usepackage{amsmath, amssymb}
\usepackage{multirow}  
\usepackage{graphicx}  
\usepackage{booktabs}  
\usepackage{makecell}
\usepackage[svgnames]{xcolor}

\usepackage[table]{xcolor}
\definecolor{tea}{RGB}{0,102,204}   
\definecolor{err}{RGB}{200,30,30} 

\newcommand{\tea}[1]{\textcolor{tea}{#1}}
\newcommand{\err}[1]{\textcolor{err}{#1}}

\usepackage{xcolor}

\definecolor{SFTBlue}{HTML}{70B7D1}
\definecolor{SFTOrange}{HTML}{F2A46F}
\definecolor{OPDGreen}{HTML}{6DD0BD}
\definecolor{OPDRed}{HTML}{F17173}

\usepackage[most]{tcolorbox}

\newtcolorbox{takeawaybox}{
    colback=white,
    colframe=black!65,
    boxrule=0.8pt,
    arc=2.5mm,
    left=3mm,
    right=3mm,
    top=2mm,
    bottom=2mm,
    before skip=6pt,
    after skip=6pt,
    width=\linewidth,
    breakable
}

\title{Simple-OPD: Demystifying Warm-up for On-policy Distillation}

\author{
Tao Liu$^{1,*}$ \quad
Taiqiang Wu$^{2,*,\dagger}$ \quad
\textbf{Mao Zheng}$^{3}$ \quad
\textbf{Xuan Luo}$^{3}$ \quad \\
\textbf{Runming Yang}$^{2}$ \quad
\textbf{Xuewei Yang}$^{1}$ \quad
\textbf{Junjie Wang}$^{1,\ddagger}$ \quad
\textbf{Yujiu Yang}$^{1,\ddagger}$
\\
$^{1}$Tsinghua University \quad
$^{2}$The University of Hong Kong \quad
$^{3}$LLM Department, Tencent
\\
\small{
$^{*}$Equal contribution \quad
$^{\dagger}$Project Leader \quad
$^{\ddagger}$Corresponding authors
}
\\
\tt\small{\href{https://github.com/Utaotao/Simple-OPD}{\faGithub\ \texttt{https://github.com/Utaotao/Simple-OPD}}}
}

\begin{document}
\maketitle
\begin{abstract}
On-policy distillation (OPD) trains a student on its own rollouts with token-level supervision from teacher models, but its effectiveness can depend strongly on the warm-up stage before OPD. 
In this paper, we demystify warm-up for OPD from both data and training perspectives. For data, we find that effective warm-up relies on teacher-compatible chain-of-thought supervision, and that even incorrect teacher rollouts can provide comparable benefits to correct ones. 
This suggests that warm-up primarily transfers a teacher-compatible thinking pattern rather than merely correct answers.
For training, we show that low-rank adaptation (LoRA) with a near-saturation training duration better balances in-domain adaptation and out-of-distribution generalization than full-parameter SFT.
Based on these findings, we propose \textbf{Simple-OPD}, a plug-and-play initialization method that warms up the student on teacher-generated CoT with LoRA before OPD. 
Experiments across diverse settings demonstrate the effectiveness and robustness of Simple-OPD.
\end{abstract}

\input{sections/1-Introduction}
\input{sections/3-Preliminaries}
\input{sections/4-warm-up_data}
\input{sections/5-Training}
\input{sections/6-More_Results}
\input{sections/2-Related}
\input{sections/7-Conclusion}

\input{main.bbl}
\input{sections/Appendix}

\end{document}

%% file: sections/1-Introduction.tex
\section{Introduction}

On-policy distillation (OPD) has emerged as an effective paradigm for transferring capabilities from one or several teacher models into a student model~\citep{lu2025onpolicydistillation, xiao2026mimo}. 
Specifically, OPD follows an on-policy paradigm that trains the student on its own sampled rollouts and employs the teacher to provide token-level dense supervision~\citep{minillm,agarwal2024policy}. 
While achieving success in transferring knowledge, it also faces the challenge that supervision from the teacher model may be biased or even harmful, especially when the student rollouts are rarely defined in the teacher's generation space.

\begin{figure}[t]
    \centering\includegraphics[width=\linewidth]{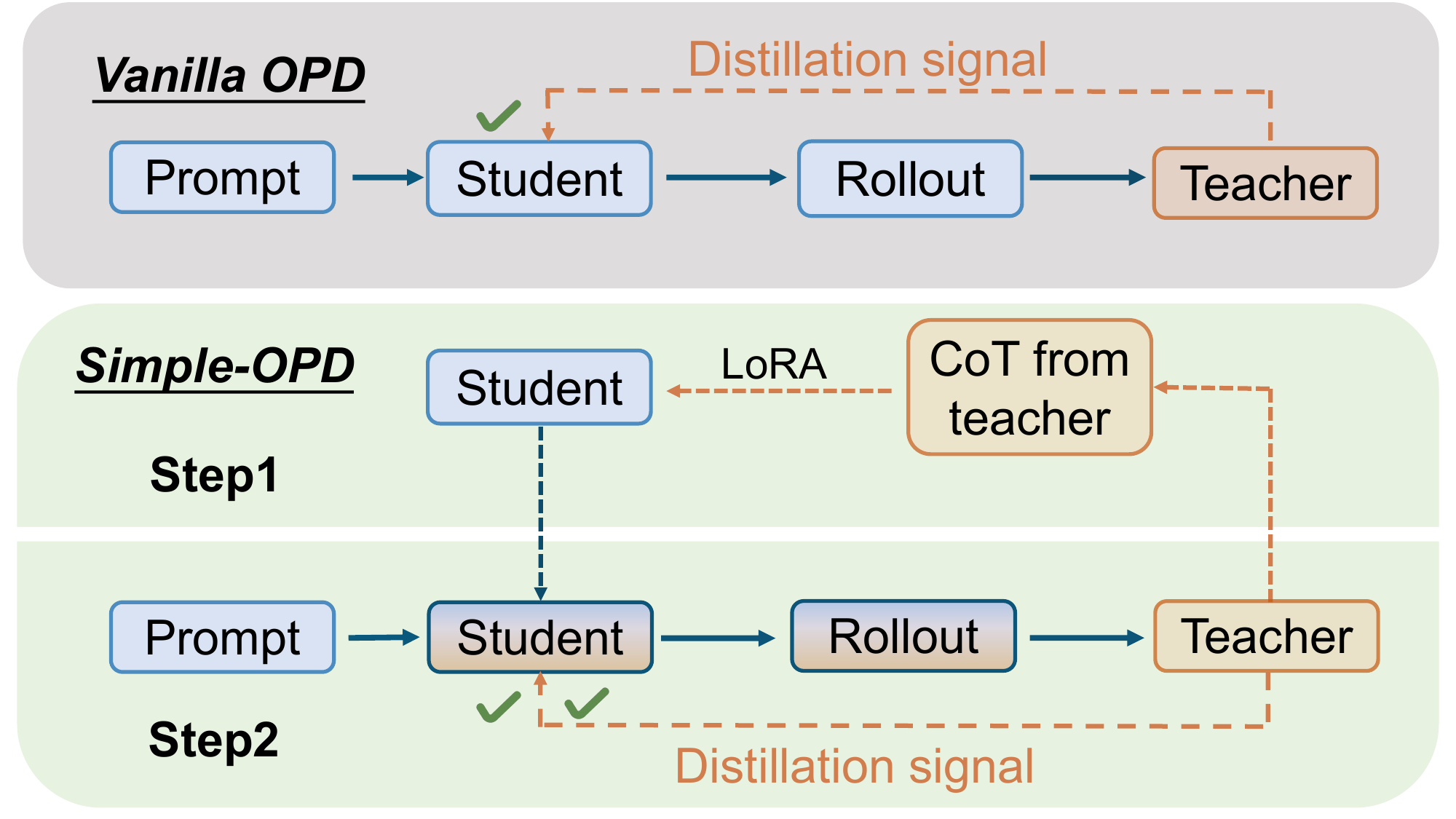}
\caption{
Overview of vanilla OPD and Simple-OPD.
Simple-OPD first warms up the student with teacher-generated CoT using LoRA, followed by standard OPD.
}
\label{fig:intro}
\end{figure}

To tackle this issue, one common practice is to conduct a warm-up for the student before OPD~\citep{xu2026deepseek, li2026rethinking}.
This stage is treated mainly as an initialization heuristic, where the student is fine-tuned on teacher responses.
In this way, the overlap between student and teacher is higher, leading to better OPD performance~\citep{li2026rethinking}.
However, the role of warm-up remains underexplored, especially the data and training recipe.

In this paper, we aim to demystify the warm-up stage for OPD and summarize practical takeaways.
For warm-up data, we find that chain-of-thought~(CoT) supervision is essential for building a strong initialization. 
For the CoT source, the OPD teacher works better than \textit{a stronger external model}~(i.e., GPT-5.5~\citep{openai2026gpt55}), suggesting that compatibility with the downstream teacher matters more than the standalone strength of the CoT generator. 
Moreover, we further observe that \textit{incorrect teacher rollouts can provide nearly the same benefit as correct ones}, indicating that warm-up transfers teacher-compatible reasoning behavior rather than simply answer correctness.

Beyond data construction, another challenge is the training recipe. 
For the training method, we find that full-parameter SFT accelerates in-domain adaptation, but degrades out-of-distribution (OOD) generalization.
Meanwhile, low-rank adaptation, LoRA~\citep{lora}, provides a better balance.
It learns the teacher-aligned reasoning pattern while limiting unnecessary changes to the student model.
We also find that warm-up duration should be approach saturation rather than be simply maximized, as overly strong warm-up may favor in-domain performance at the cost of generalization.

Based on these findings, we propose Simple-OPD, a simple-yet-effective and plug-and-play OPD initialization method. 
As illustrated in Figure~\ref{fig:intro}, we warm up the student with CoT rollouts generated by the OPD teacher using a sufficiently trained low-rank LoRA adapter.
Across different OPD objectives, thinking and non-thinking model settings, and same-size teacher-student consolidation, Simple-OPD consistently improves in-domain reasoning performance while preserving overall out-of-domain generalization.
Our contributions are summarized as follows:
  \begin{itemize}
      \item We systematically investigate the warm-up stage for OPD and summarize practical takeaways regarding data and training recipe.
      \item We propose a simple-yet-effective Simple-OPD method, employing LoRA on the teacher's rollouts for warm-up. 
      \item We conduct experiments across various settings to demonstrate the effectiveness and robustness of the proposed Simple-OPD.  
  \end{itemize}
  

%% file: sections/3-Preliminaries.tex
\section{Preliminaries}
\label{sec:preliminaries}

\paragraph{Supervised Fine-Tuning(SFT).}
Let $\pi_{\theta}$ denote the student policy with parameters
$\theta$. Let $\mathcal{D}_{\mathrm{SFT}}$ consist of pairs
$(x,y^{*})$, where $x$ is a prompt and
$y^{*}=(y^{*}_{1},\ldots,y^{*}_{T})$ is a reference response
of length $T$. We use $y^{*}_{<t}$ to denote the response
prefix before position $t$, and define the token loss as
$\ell^{\mathrm{SFT}}_{t}(\theta)
= -\log \pi_{\theta}(y^{*}_{t}\mid x,y^{*}_{<t})$.
The SFT objective is
\begin{equation}
\mathcal{L}_{\mathrm{SFT}}(\theta)
=
\mathbb{E}_{(x,y^{*})\sim\mathcal{D}_{\mathrm{SFT}}}
\left[
\frac{1}{T}
\sum_{t=1}^{T}
\ell^{\mathrm{SFT}}_{t}(\theta)
\right].
\label{eq:sft}
\end{equation}
We use SFT to initialize the student before OPD and refer to
this stage as warm-up.

\paragraph{On-Policy Distillation(OPD).}
Let $\pi_{\mathrm{T}}$ denote the fixed teacher policy and
$\mathcal{D}_{x}$ the prompt distribution. Given a prompt
$x\sim\mathcal{D}_{x}$, the student samples a response
$\hat{y}=(\hat{y}_{1},\ldots,\hat{y}_{L})$ from
$\pi_{\theta}(\cdot\mid x)$, where $L=|\hat{y}|$.
At position $t$, the context is
$c_t=(x,\hat{y}_{<t})$. We define the token reverse KL as
\[
\ell^{\mathrm{RKL}}_{t}(\theta)
=
D_{\mathrm{KL}}
\left(
\pi_{\theta}(\cdot\mid c_t)
\Vert
\pi_{\mathrm{T}}(\cdot\mid c_t)
\right).
\]
OPD minimizes its average over student rollouts:
\begin{equation}
\mathcal{L}_{\mathrm{OPD}}(\theta)
=
\mathbb{E}_{\substack{
x\sim\mathcal{D}_{x}\\
\hat{y}\sim\pi_{\theta}(\cdot\mid x)
}}
\left[
\frac{1}{L}
\sum_{t=1}^{L}
\ell^{\mathrm{RKL}}_{t}(\theta)
\right].
\label{eq:opd}
\end{equation}
Since $\hat{y}$ is sampled from $\pi_{\theta}$, the contexts
in the objective are induced by the current student policy.

\paragraph{Low-Rank Adaptation (LoRA).}
Let $W_{0}\in\mathbb{R}^{d\times k}$ denote a frozen weight matrix with input dimension $k$ and output dimension $d$.
LoRA introduces trainable matrices $A\in\mathbb{R}^{r\times k}$ and $B\in\mathbb{R}^{d\times r}$, where $r\ll\min(d,k)$ is the adaptation rank\citep{lora,wu2024mixture}. 
Given a scaling coefficient $\alpha$, the adapted weight matrix is
\begin{equation}
W
=
W_{0}
+
\frac{\alpha}{r}BA.
\label{eq:lora}
\end{equation}
During adaptation, $W_{0}$ remains frozen while $A$ and $B$
are optimized.

%% file: sections/4-warm-up_data.tex
\begin{figure}[t]
    \centering
    \includegraphics[width=\linewidth]{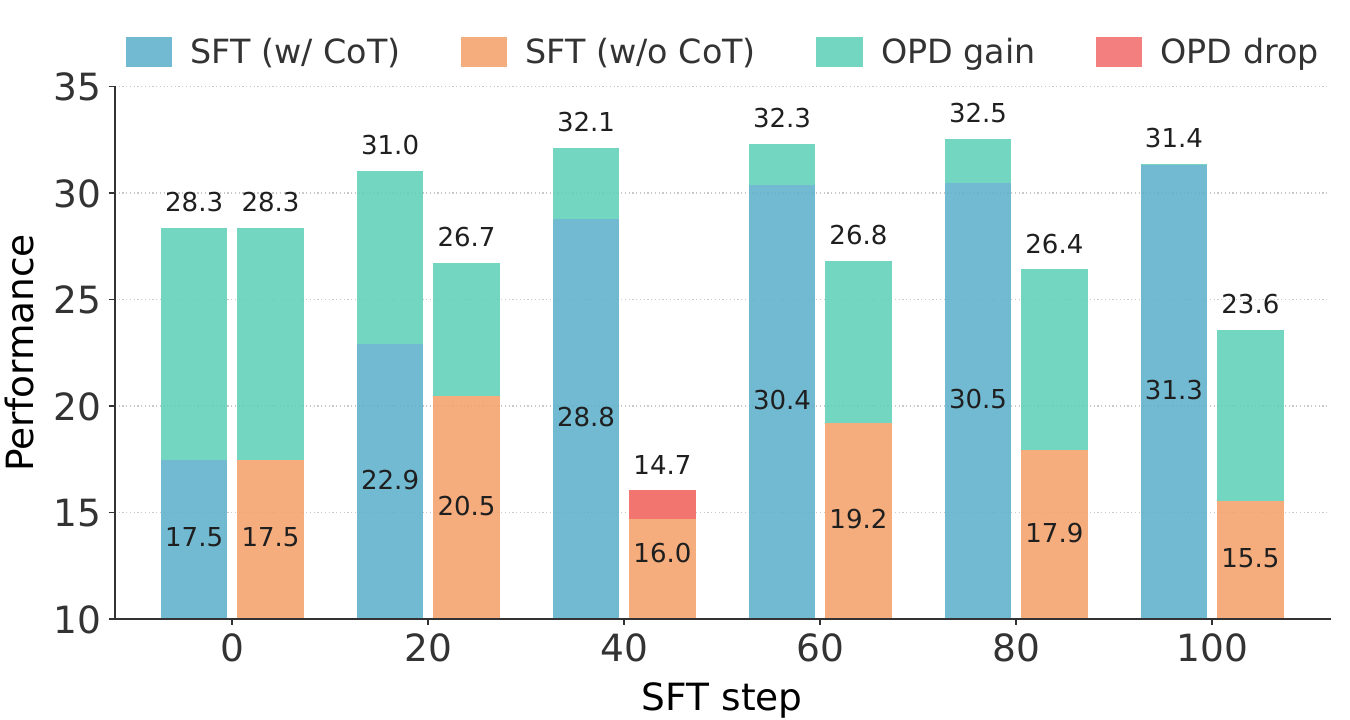}
\caption{
Effect of CoT supervision during warm-up.
The \textcolor{SFTBlue}{blue} and \textcolor{SFTOrange}{orange} bars denote SFT with and without CoT, respectively.
The \textcolor{OPDGreen}{green} and \textcolor{OPDRed}{red} segments denote the gain and drop after OPD.
CoT provides a stronger and more stable initialization for OPD.
}
    \label{fig:cot_gap}
\end{figure}

\section{Data for Warm-up}
\label{sec:data}
Warm-up provides the initialization for subsequent OPD, making the construction of its training data particularly important.
In this section, we study three properties of warm-up data: whether they contain CoT, whether the CoT is generated by the OPD teacher, and whether the rollouts are correct.
\paragraph{Experimental Setup.}
We use Qwen3-1.7B-Base as the student, and Qwen3-8B-Base trained on DAPO-Math-17K~\citep{yu2026dapo} as the teacher~\citep{yang2026internalize}.
DAPO-Math-17K is also used for subsequent OPD, and the prompts for SFT warm-up are selected from this dataset.
Each warm-up checkpoint is followed by 75 steps of OPD.
Unless otherwise specified, we report the average performance on MATH-500~\citep{hendrycks2021measuring}, AIME24~\citep{aime2024}, and AIME25~\citep{aime2025}.
We use $\mathrm{avg@4}$ for MATH-500 and $\mathrm{avg@16}$ for AIME24 and AIME25.
We employ verl~\citep{sheng2024hybridflow} as the primary training framework and leverage
vllm~\citep{kwon2023efficient} to speed up the sampling process. 
All experiments are conducted on eight GPUs.
Detailed data construction, training hyperparameters, and evaluation inference settings are provided in Appendix~\ref{app:hyper-parameters}.



\subsection{CoT is Essential for Warm-up}
\label{sec:cot_presence}
We compare SFT warm-up using rollouts with CoT and rollouts containing only final answers.
As shown in Figure~\ref{fig:cot_gap}, step 0 corresponds to direct OPD from the base student without SFT warm-up.

\textit{CoT provides a sustained learning signal during warm-up.}
At the same training step, checkpoints trained with CoT consistently outperform those trained without CoT, and the gap generally widens as warm-up proceeds.
This indicates that exposing the reasoning process provides more effective supervision than learning from final answers alone.

\textit{The advantage of CoT warm-up persists after subsequent OPD.}
Although OPD can substantially improve several no-CoT checkpoints, it does not compensate for their weaker initialization.
Across all matched steps, CoT warm-up followed by OPD consistently achieves stronger final performance.

\textit{Sufficient CoT warm-up makes subsequent OPD more stable and less dependent on checkpoint selection.}
As more reasoning capability is acquired during warm-up, the additional benefit from OPD generally becomes smaller.
Meanwhile, CoT checkpoints converge to similar final results, whereas no-CoT checkpoints exhibit much larger fluctuations.
These findings show that CoT is essential for building a strong and reliable initialization for OPD.
Detailed results on each benchmark are reported in Appendix~\ref{app:cot_presence}.

\begin{figure}[t]
    \centering
    \includegraphics[width=\linewidth]{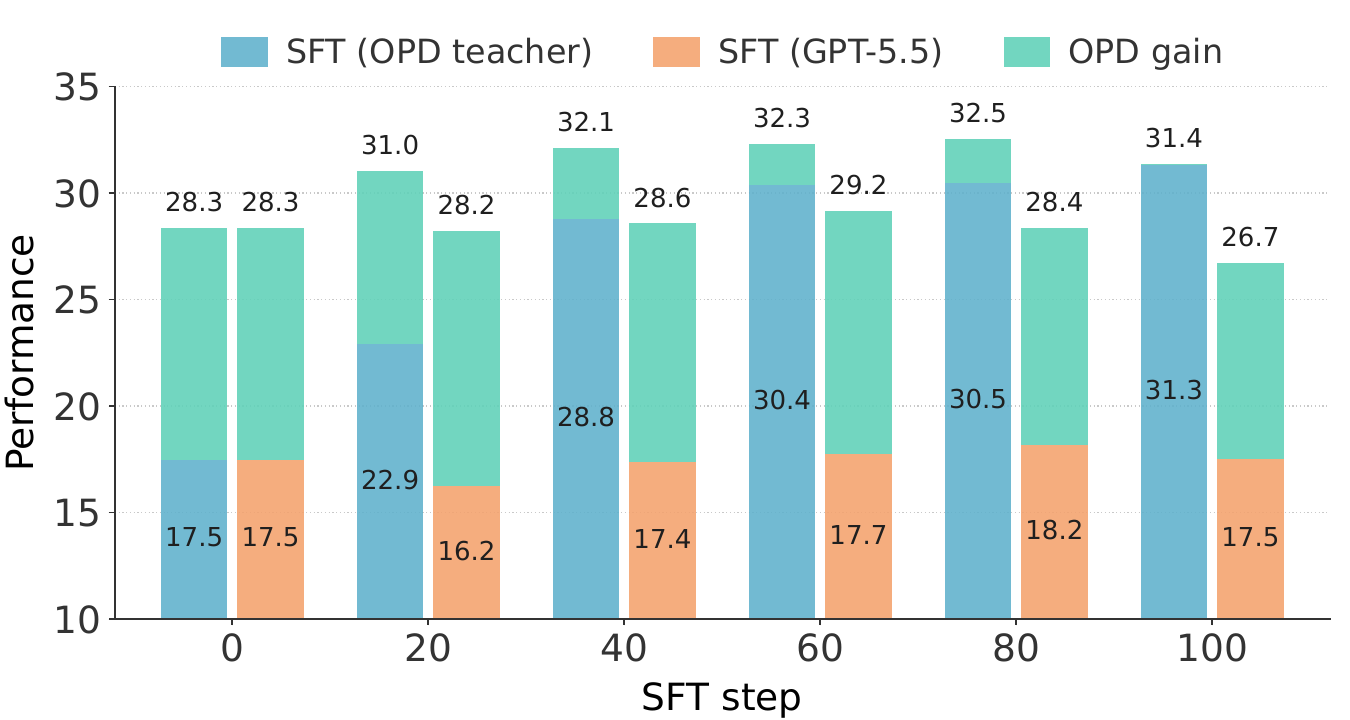}
\caption{
Performance comparison of CoT source, i.e., OPD teacher and GPT-5.5.
OPD teacher consistently outperforms GPT-5.5 for SFT and subsequent OPD.
This suggests that effective warm-up depends on alignment with the downstream teacher rather than reasoning quality alone.
}
    \label{fig:cot_source}
\end{figure}

\subsection{Teacher's CoT Works Better}
\label{sec:cot_source}

We next examine whether CoT from a stronger external model can also provide an effective warm-up for OPD.
Using the same prompts, we construct warm-up rollouts with either the OPD teacher or GPT-5.5.
Although GPT-5.5 is substantially stronger than the OPD teacher, its CoT produces a much weaker initialization.

\textit{Only CoT generated by the OPD teacher consistently improves warm-up performance.}
As shown in Figure~\ref{fig:cot_source}, teacher-generated CoT steadily improves the student as warm-up proceeds.
In contrast, GPT-5.5 CoT leaves the student close to its initial performance despite being generated by a stronger model.

\textit{This difference persists after subsequent OPD.}
Across all warm-up durations, checkpoints trained on the teacher's CoT outperform both GPT-5.5 warm-up and direct OPD.
Although OPD partially recovers the weak performance following GPT-5.5 warm-up, the final results remain close to the direct OPD baseline and deteriorate with longer warm-up.

These results show that the effectiveness of warm-up is determined not simply by the strength of the CoT generator.
Instead, warm-up requires CoT that is compatible with the teacher used during subsequent OPD.
A benchmark-level comparison across the two CoT sources is provided in Appendix~\ref{app:cot_source}.

\subsection{Wrong Rollout Also Works}
\label{sec:cot_correctness}

\begin{figure}[t]
    \centering
    \includegraphics[width=\linewidth]{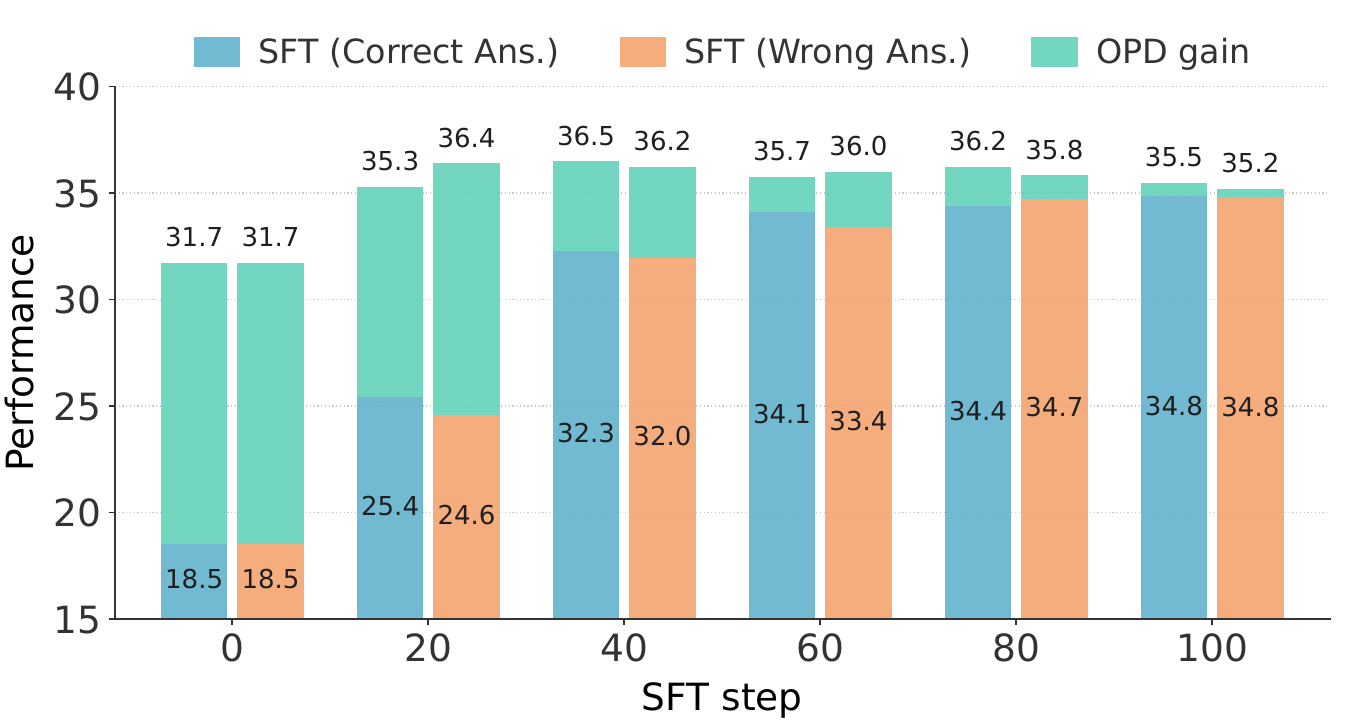}
\caption{
Effect of correctness for teacher's rollouts.
Correct and wrong rollouts produce similar performance after both SFT and subsequent OPD.
The key is to expose the teacher's thinking patterns rather than answers.
}
    \label{fig:cot_correctness}
\end{figure}

Having established that warm-up requires CoT aligned with the OPD teacher, we next examine whether these rollouts must also be correct.
We construct paired correct and wrong rollouts from the OPD teacher for the same prompts, so that both settings contain CoT from the same teacher.
For this ablation, we additionally include AMC23 \citep{amc2023} and report the average over four benchmarks.
As shown in Figure~\ref{fig:cot_correctness}, the two settings produce similar SFT trajectories and differ by less than one point at most checkpoints.

The same pattern remains after OPD.
The final scores of both settings stay within a narrow range of 35.2 to 36.5, and neither condition shows a consistent advantage.
Their relative performance after SFT also does not consistently carry over to OPD, showing that a stronger SFT checkpoint does not necessarily provide a better initialization.
These results suggest that once CoT from the OPD teacher is available, rollout correctness has only a limited effect on subsequent OPD.
Detailed benchmark-level results and an example pair of correct and incorrect rollouts are provided in Appendix~\ref{app:cot_correctness}.

\begin{takeawaybox}
\textbf{Takeaway on Warm-up Data.}
\textit{
Together, the three ablations show that effective warm-up requires CoT aligned with the OPD teacher, while rollout correctness plays only a secondary role.
This suggests that warm-up primarily transfers a teacher-compatible thinking pattern rather than merely correct answers.
}
\end{takeawaybox}

%% file: sections/5-Training.tex
\begin{figure*}[t]
    \centering
    \includegraphics[width=\textwidth]{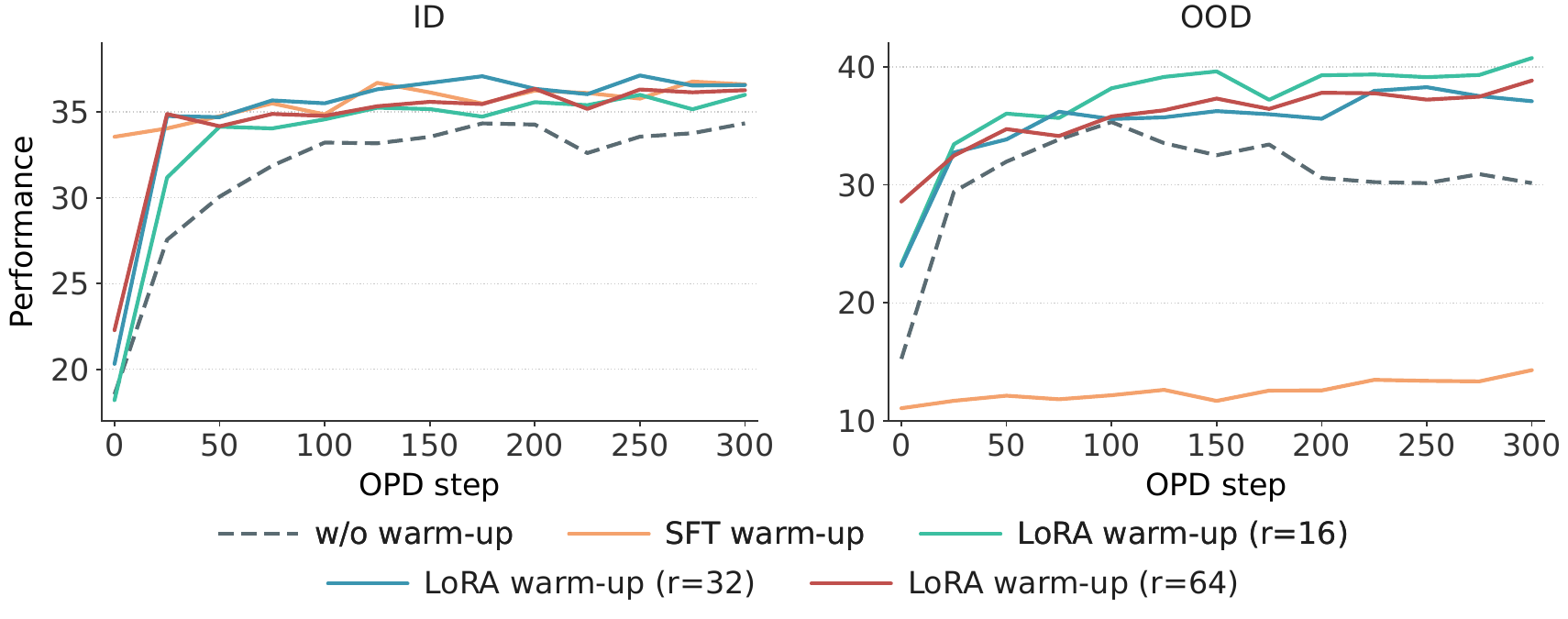}
    \caption{
        Comparison of full SFT and LoRA warm-up during OPD.
        Full SFT accelerates in-domain convergence but causes persistent OOD degradation.
        LoRA reaches comparable in-domain performance while retaining substantially stronger generalization.
    }
    \label{fig:lora_warm-up}
\end{figure*}

\section{Training Recipe for Warm-up}
\label{sec:training}
Having identified the key properties of warm-up data, we next study how warm-up should be trained.
Since effective warm-up should facilitate in-domain adaptation without sacrificing general capabilities, we evaluate both ID performance and OOD generalization when examining its parameterization and training strength.

\paragraph{Experimental Setup.}
We follow the student and teacher setup described in Section~\ref{sec:data}.
We report the average performance over AMC23, MATH-500, AIME24, and AIME25 as the in-domain result.
For out-of-domain evaluation, we use IFEval~\citep{zhou2023instruction}, GPQA-Diamond~\citep{rein2024gpqa}, HumanEval~\citep{chen2021codex}, and the Chemistry, Physics, and History subsets of MMLU-Pro~\citep{wang2024mmlu}.
For the out-of-domain evaluation pipeline, we utilized OpenCompass~\citep{2023opencompass} as the primary inference framework, integrated with LMDeploy~\citep{2023lmdeploy} as the acceleration backend. 
We report the average performance over the six out-of-domain benchmarks as the overall OOD result.

\subsection{LoRA Warm-up Balances Adaptation and Generalization}
\label{sec:lora_warmup}

We compare direct OPD with full-parameter SFT warm-up and LoRA warm-up at ranks 16, 32, and 64.
Figure~\ref{fig:lora_warm-up} reports their ID and OOD trajectories during subsequent OPD.

\textit{Direct OPD and full SFT warm-up exhibit different limitations.}
Direct OPD gradually improves ID performance, but its OOD performance peaks early and then declines as training continues.
Full SFT warm-up substantially raises the initial ID performance and accelerates convergence.
However, it severely reduces OOD performance before OPD, and subsequent training fails to recover the lost generalization.

\textit{
LoRA warm-up provides a better balance between ID adaptation and OOD generalization.
}
Although LoRA checkpoints initially underperform full SFT on ID benchmarks, they improve rapidly and eventually reach comparable performance.
On OOD benchmarks, all LoRA variants start above both the base student and the full SFT checkpoint.
This advantage is maintained or further enlarged during OPD.
Therefore, LoRA preserves the ID benefit of warm-up while avoiding the severe OOD degradation caused by full-parameter adaptation.

\textit{A relatively small LoRA rank is already sufficient for effective warm-up.}
Ranks 16, 32, and 64 achieve similar final ID performance, indicating that increasing the adaptation capacity provides little additional ID benefit.
In contrast, rank 16 produces the strongest OOD trajectory, while ranks 32 and 64 show slightly weaker generalization.
This pattern suggests that a more constrained update can learn the target reasoning behavior while introducing less interference with the pretrained capabilities.

Overall, LoRA acts as both a parameter-efficient adaptation method and a constraint on warm-up updates.
A low-rank update is sufficient to acquire the teacher-aligned thinking pattern while limiting unnecessary changes to the pretrained model.
It therefore offers a more favorable balance between adaptation and generalization than full-parameter warm-up.

\subsection{A Moderate Warm-up Step is Preferred}
\label{sec:lora_steps}

\begin{figure*}[t]
    \centering
    \includegraphics[width=\textwidth]{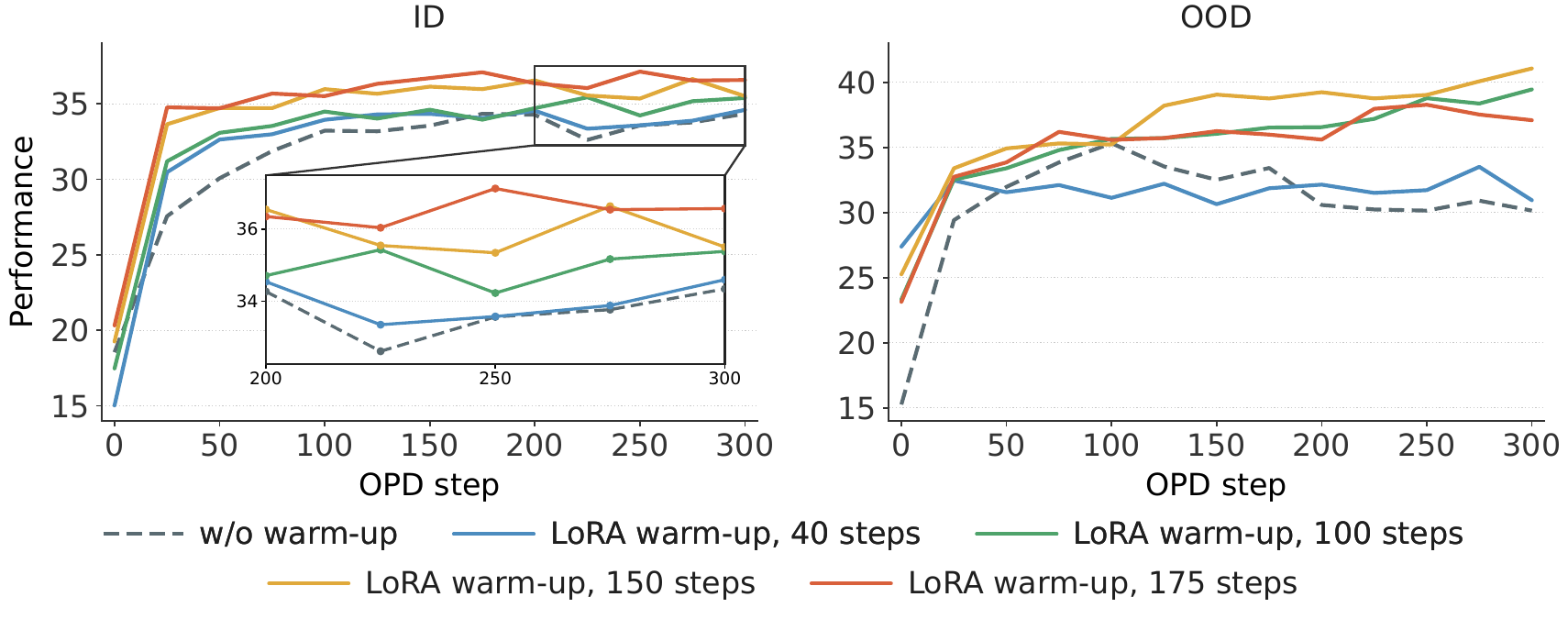}
    \caption{
        Effect of warm-up duration with the LoRA rank fixed to 32.
        A short warm-up remains close to direct OPD, while longer warm-up improves both performance and convergence speed.
        The 150-step setting provides the best ID--OOD balance, whereas 175 steps slightly favor ID performance.
    }
    \label{fig:lora_steps}
\end{figure*}

We study the effect of warm-up steps while fixing the LoRA rank to 32.
Figure~\ref{fig:lora_steps} compares warm-up checkpoints obtained after 40, 100, 150, and 175 steps.

\textit{Insufficient LoRA warm-up provides only limited benefits.}
With 40 warm-up steps, the ID trajectory remains close to direct OPD, while the OOD performance shows no clear advantage as training proceeds.
Increasing the duration to 100 steps yields a modest ID improvement and a more substantial gain in OOD generalization.

\textit{Sufficiently trained LoRA warm-up improves both ID and OOD performance, but the gains are not monotonic with training duration.}
The 150-step and 175-step settings consistently outperform shorter warm-up configurations in both domains.
The 175-step setting performs slightly better on ID benchmarks, whereas the 150-step setting achieves the strongest OOD performance.
Therefore, the preferred checkpoint lies near saturation rather than simply at the maximum training duration.

\textit{LoRA warm-up also accelerates subsequent OPD.}
The sufficiently trained warm-up variants approach their stable ID performance after roughly 100 OPD steps, whereas direct OPD continues improving until around 200 steps.
This is consistent with the training dynamics in Figure~\ref{fig:training-dynamics}, where warm-up leads to faster and more stable optimization.
Warm-up therefore reduces the number of OPD updates required to reach competitive performance.
Additional comparisons across LoRA ranks and warm-up steps are provided in Appendix~\ref{app:lora-rank-warm-up}.

\begin{takeawaybox}
\textbf{Takeaway on Warm-up Training.}
\textit{
For warm-up training, we recommend a relatively low-rank LoRA trained to near saturation.
This configuration provides a Pareto-efficient balance between ID performance and OOD generalization while accelerating subsequent OPD.
}
\end{takeawaybox}

\subsection{Proposed Simple-OPD}
\label{sec:simple_opd}

Based on the findings in Sections~\ref{sec:data} and~\ref{sec:training}, we propose \textbf{Simple-OPD}.
It uses CoT rollouts generated by the OPD teacher to warm up the student with a sufficiently trained low-rank LoRA adapter, followed by standard OPD.
This simple recipe provides a stronger initialization, accelerates subsequent training, and better preserves OOD generalization without modifying the OPD objective.

%% file: sections/6-More_Results.tex
\section{Extensive Analysis}
\label{sec:further_results}

\input{tables/opd_variants}

\subsection{Compatibility with OPD Variants}
\label{sec:opd_variants}

We examine whether Simple-OPD can complement different OPD objectives rather than being tied to vanilla OPD.
We consider standard OPD, G-OPD~\citep{yang2026gopd}, and PowerOPD~\citep{zhao2026poweropd}.
We use Qwen3-1.7B in non-thinking mode as the student and Qwen3-4B in non-thinking mode as the teacher~\citep{yang2025qwen3}.
The teacher is trained with GRPO~\citep{shao2024deepseekmath} on DAPO-Math-17K~\citep{yu2026dapo}.
For each objective, we compare direct training from the base student with its Simple-OPD counterpart initialized by teacher-CoT LoRA warm-up.
The ID score is averaged over AIME24 and AIME25.
The OOD score is averaged over IFEval, GPQA-Diamond, and the Physics, Chemistry, and History subsets of MMLU-Pro.

\textit{Simple-OPD consistently improves ID performance across different OPD objectives.}
As shown in Table~\ref{tab:opd_variants}, it improves the ID average by 1.35 points for vanilla OPD, 1.57 points for G-OPD, and 0.95 points for PowerOPD.
Its benefit therefore persists when the underlying distillation objective changes.

\textit{Simple-OPD preserves overall OOD performance while producing method-dependent changes on individual benchmarks.}
The average OOD score improves for vanilla OPD and PowerOPD and remains nearly unchanged for G-OPD.
The gains appear on different benchmarks under different objectives, indicating that the effect depends partly on the subsequent OPD method.
Overall, Simple-OPD strengthens ID performance without introducing systematic OOD degradation.

These results show that Simple-OPD is complementary to the choice of OPD objective and can serve as a general initialization recipe for different OPD variants.

\input{tables/thinking_mode}

\subsection{Applied to Thinking Models}
\label{sec:thinking_mode}

We further evaluate Simple-OPD on thinking models.
We use Qwen3-0.6B with thinking enabled as the student and Qwen3-4B-Thinking-2507 as the teacher~\citep{yang2025qwen3}.
We compare direct OPD with Simple-OPD under the same training configuration.

\textit{Simple-OPD consistently improves mathematical reasoning in the thinking setting.}
As shown in Table~\ref{tab:thinking_mode}, direct OPD raises the ID average from 37.94 to 42.36, while Simple-OPD further improves it to 43.81.
The gains mainly come from AMC23, MATH-500, and AIME25, with only a slight decrease on AIME24.

\textit{Simple-OPD also improves overall OOD generalization.}
The OOD average increases from 35.43 to 36.28, with improvements on IFEval, GPQA-Diamond, and MMLU-Pro History.
Physics remains nearly unchanged, while Chemistry decreases moderately.
Overall, these results show that Simple-OPD remains effective when both the student and teacher operate in thinking mode.

\subsection{OPD for a Student-Trained Teacher}
\label{sec:same_size}

We further evaluate Simple-OPD for same-size model consolidation.
We use DeepSeek-R1-Distill-Qwen-1.5B as the student and the RL-trained JustRL-DeepSeek-1.5B as the teacher~\citep{guo2025deepseek,he2025justrl}.
This setting reflects a practical scenario in which capabilities acquired by a stronger post-trained model are consolidated into a deployment model of the same size.

\begin{figure}[t]
    \centering
    \includegraphics[width=\linewidth]{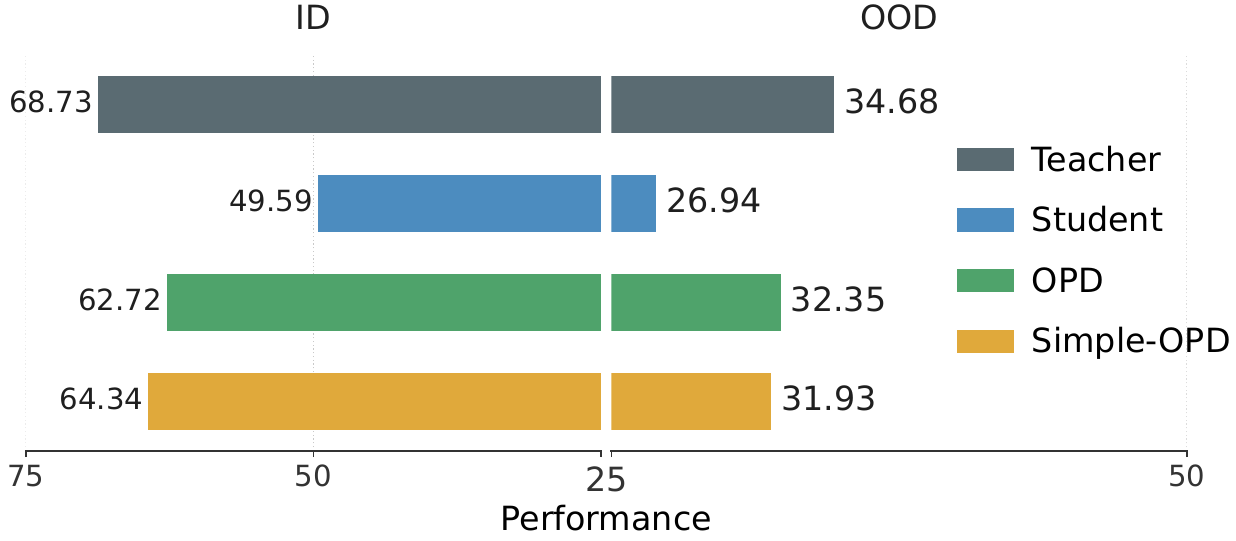}
\caption{
Performance with a same-size student-trained teacher.
Simple-OPD further improves the ID performance of direct OPD while keeping the overall OOD performance nearly unchanged.
}
\label{fig:same_size}
\end{figure}

\textit{Simple-OPD substantially improves ID performance while largely preserving OOD generalization.}
As shown in Figure~\ref{fig:same_size}, direct OPD raises the ID average from 49.59 to 62.72, and Simple-OPD further improves it to 64.34.
Meanwhile, the OOD average changes only slightly from 32.35 to 31.93.
This result shows that Simple-OPD effectively consolidates the capabilities of a same-size post-trained teacher with only a marginal OOD trade-off.

Together with the preceding Qwen3 experiments, these results further demonstrate the effectiveness of Simple-OPD across different model architectures.
Detailed benchmark-level results are provided in Table~\ref{tab:same_size}.

\subsection{Training Dynamics}
\begin{figure*}[t]
    \centering
    \includegraphics[width=\textwidth]{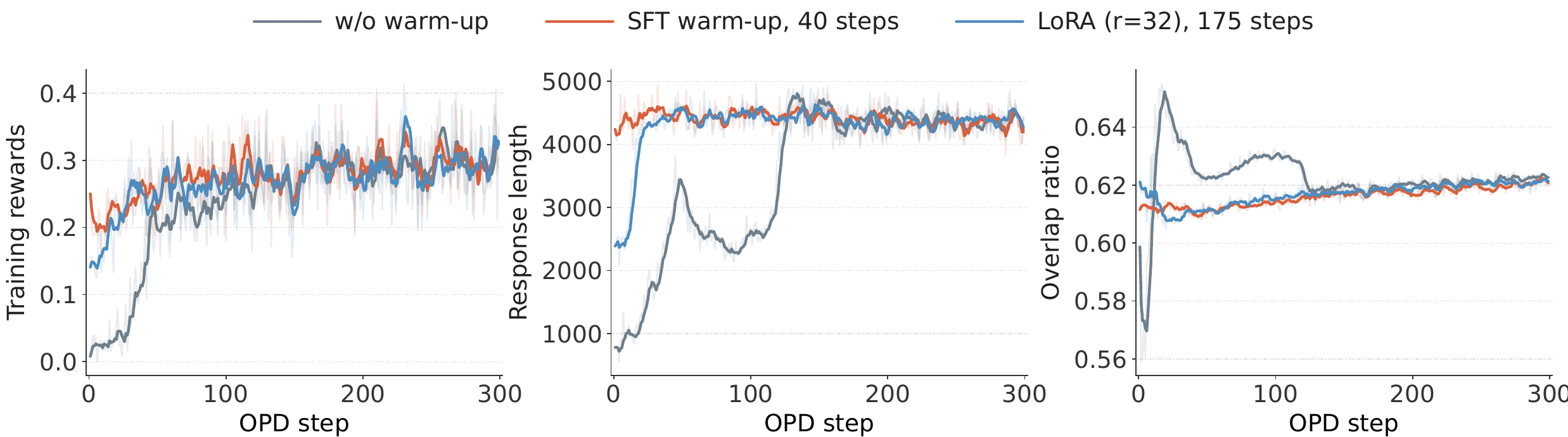}
    \caption{
    OPD training dynamics with and without SFT warm-up.
    From left to right, we report training reward, response length, and overlap ratio over OPD steps.
    Compared with training from the base model directly, both full-parameter and LoRA warm-up lead to faster and more stable convergence during the early stage of OPD.
    }
    \label{fig:training-dynamics}
\end{figure*}

Figure~\ref{fig:training-dynamics} compares the OPD training dynamics with and without SFT warm-up.
The warm-up models start with higher training rewards and reach a stable response-length regime substantially earlier than the no-warm-up model.
Warm-up also reduces the large early-stage fluctuations in the overlap ratio(defined in Appendix~\ref{app:overlap_ratio}).
After sufficient OPD training, the different configurations gradually approach similar regimes.
These results suggest that SFT warm-up primarily accelerates and stabilizes the convergence of subsequent OPD training.

\subsection{Case Study}
\label{sec:case_stu}

To better understand how warm-up affects the student's reasoning behavior, we examine representative MATH-500 cases at OPD step~\(300\).
As shown in Table~\ref{tab:case_pattern} in the Appendix, the warm-up model exhibits a reasoning pattern closer to that of the teacher.
Both the teacher and the warm-up model explicitly verify intermediate results or independently recompute the solution before producing the final answer.
In contrast, the no-warm-up model follows a plausible solution path but terminates without verification, allowing local counting or arithmetic errors to remain undetected.
These examples suggest that warm-up transfers teacher-like verification behavior, thereby improving the reliability of subsequent rollouts.

%% file: tables/opd_variants.tex
\begin{table*}[t]
\centering
\begin{tabular}{lccc@{\hspace{6pt}}cccccc}
\toprule
\textbf{Method}
& \multicolumn{3}{c}{\textbf{ID}}
& \multicolumn{6}{c}{\textbf{OOD}} \\
\cmidrule(lr){2-4}
\cmidrule(lr){5-10}
& \textbf{AIME24}
& \textbf{AIME25}
& \textbf{Avg.}
& \textbf{IFEval}
& \textbf{GPQA}
& \textbf{Phys.}
& \textbf{Chem.}
& \textbf{Hist.}
& \textbf{Avg.} \\
\midrule

Teacher
& 63.75
& 56.25
& 60.00
& 80.96
& 43.43
& 78.21
& 77.47
& 48.29
& 65.67 \\
\midrule

Student
& 13.75
& 11.87
& 12.81
& 65.80
& 28.79
& 42.03
& 40.46
& 25.46
& 40.51 \\
\midrule

OPD
& 40.21
& 36.46
& 38.34
& \textbf{64.88}
& 20.71
& 57.89
& \textbf{60.60}
& \textbf{33.07}
& 47.43 \\
\rowcolor{blue!5}
Simple-OPD
& \textbf{42.50}
& \textbf{36.88}
& \textbf{39.69}
& 63.96
& \textbf{26.77}
& \textbf{60.51}
& 60.51
& \textbf{33.07}
& \textbf{48.96} \\
\addlinespace

G-OPD
& 44.79
& 38.33
& 41.56
& \textbf{64.51}
& \textbf{13.64}
& 54.04
& 55.65
& \textbf{35.70}
& \textbf{44.71} \\
\rowcolor{blue!5}
Simple-OPD
& \textbf{46.88}
& \textbf{39.37}
& \textbf{43.13}
& 63.96
& 13.13
& \textbf{55.74}
& \textbf{57.69}
& 32.55
& 44.61 \\
\addlinespace

PowerOPD
& \textbf{43.54}
& 34.58
& 39.06
& 62.66
& \textbf{21.72}
& \textbf{59.74}
& \textbf{59.98}
& 29.92
& 46.80 \\
\rowcolor{blue!5}
Simple-OPD
& 43.13
& \textbf{36.88}
& \textbf{40.01}
& \textbf{65.80}
& 20.20
& 57.97
& 59.01
& \textbf{35.17}
& \textbf{47.63} \\
\bottomrule
\end{tabular}

\caption{
Compatibility of Simple-OPD with different OPD variants.
We use Qwen3-1.7B (non-thinking) and a GRPO-trained Qwen3-4B (non-thinking) as the student and teacher, respectively.
Simple-OPD consistently improves ID performance while preserving overall OOD performance across three OPD objectives.
Within each OPD setting, the better result is shown in bold.
}
\label{tab:opd_variants}
\end{table*}

%% file: tables/thinking_mode.tex
\begin{table*}[t]
\centering
\setlength{\tabcolsep}{3.0pt}
\renewcommand{\arraystretch}{1.12}
\begin{tabular}{@{}lccccc@{\hspace{6pt}}cccccc@{}}
\toprule
\textbf{Method}
& \multicolumn{5}{c}{\textbf{ID}}
& \multicolumn{6}{c}{\textbf{OOD}} \\
\cmidrule(lr){2-6}
\cmidrule(lr){7-12}
& \makecell{\textbf{AIME}\\\textbf{24}}
& \makecell{\textbf{AIME}\\\textbf{25}}
& \makecell{\textbf{AMC}\\\textbf{23}}
& \makecell{\textbf{MATH}\\\textbf{500}}
& \textbf{Avg.}
& \textbf{IFEval}
& \textbf{GPQA}
& \textbf{Phys.}
& \textbf{Chem.}
& \textbf{Hist.}
& \textbf{Avg.} \\
\midrule

Teacher
& 76.25
& 75.00
& 99.69
& 97.70
& 87.16
& 30.50
& 66.67
& 83.91
& 84.72
& 58.79
& 64.92 \\
\midrule

Student
& 9.17
& 18.75
& 49.69
& 74.15
& 37.94
& 33.09
& 22.22
& 43.42
& 45.49
& 17.85
& 32.41 \\
\midrule

OPD
& \textbf{14.17}
& 22.08
& 54.37
& 78.80
& 42.36
& 29.76
& 24.24
& \textbf{50.73}
& \textbf{52.21}
& 20.21
& 35.43 \\
\rowcolor{blue!5}
Simple-OPD
& 12.92
& \textbf{23.33}
& \textbf{57.19}
& \textbf{81.80}
& \textbf{43.81}
& \textbf{30.87}
& \textbf{26.77}
& 50.65
& 50.80
& \textbf{22.31}
& \textbf{36.28} \\
\bottomrule
\end{tabular}

\caption{
Results in the Qwen3 thinking setting.
We use Qwen3-0.6B with thinking enabled as the student and Qwen3-4B-Thinking-2507 as the teacher.
Simple-OPD improves both the ID and OOD averages over direct OPD.
The Simple-OPD row is shaded, and the better result within each OPD comparison is shown in bold.
}
\label{tab:thinking_mode}
\end{table*}

%% file: sections/2-Related.tex
\textbf{}\section{Related Work}

\paragraph{On-Policy Distillation.}
OPD trains a student on its own rollouts using token-level supervision from a teacher \citep{lu2025onpolicydistillation}. MiniLLM applies reverse KL to this setting, while GKD extends the framework to multiple divergence objectives and mixtures of student and teacher trajectories \citep{minillm,agarwal2024policy}. 
Recent work has improved OPD through generalized reward and reference formulations \citep{yang2026gopd,feng2026weak}, uncertainty aware and asymmetric objectives \citep{jin2026entropy,jia2026asymmetric,ko2026scaling}, and more efficient rollout construction \citep{zhang2026prefix,yang2026prune,wu2026lightning,ziheng2026less}. 
Other studies reduce gradient variance and improve training stability \citep{oh2026kl,luo2026demystifying,jang2026stable}, or analyze the failure modes and effective conditions of OPD \citep{fu2026revisiting,li2026rethinking,kim2026does,kaur2026rethinking}. 
These studies mainly focus on the distillation objective and rollout process, while the supervised warm-up before OPD remains less explored.

\paragraph{Warm-up for OPD.}
Qwen3 adopts a two-stage strong-to-weak distillation pipeline, performing off-policy response distillation before OPD \citep{yang2025qwen3}. \citet{li2026rethinking} introduce an off-policy cold start to align student and teacher reasoning patterns. 
\citet{xu2026sparse} apply forward KL warm-up, while \citet{wu2026lightning} emphasize teacher consistency across SFT and OPD. 
Other methods improve early rollouts through behavior blending or phased teacher sampling \citep{plyusov2026trust,xu2026sgopd,luo2026demystifying}. 
These studies mainly use warm-up as a fixed initialization or stabilization stage.
In this work, we systematically study its data construction, training strength, and parameterization, and evaluate how these choices affect both ID performance and OOD generalization.


%% file: sections/7-Conclusion.tex
\section{Conclusion}

In this work, we systematically investigated the warm-up stage for on-policy distillation from both data and training perspectives. 
For data construction, the OPD teacher's CoT works better than CoT from a stronger external model, while incorrect teacher rollouts can still provide comparable benefits. 
For training, a near-saturation low-rank LoRA warm-up offers a better balance between ID and OOD domains.
Based on these findings, we propose Simple-OPD, a plug-and-play initialization recipe that warms up the student with teacher-generated CoT before OPD. 
Experiments across different OPD objectives, model settings, and teacher-student configurations show that Simple-OPD consistently improves in-domain reasoning while preserving overall generalization.


\section*{Limitation}

This work focuses on understanding and improving the warm-up stage for OPD through controlled experiments under representative model and benchmark settings. 
While our experiments cover multiple OPD objectives, thinking and non-thinking models, and same-size teacher-student consolidation, broader validation on additional model families, domains, and larger-scale training settings would further strengthen the empirical conclusions.
In addition, Simple-OPD is intentionally designed as a simple initialization recipe without modifying the OPD objective.
Future work may explore how warm-up design interacts with more advanced objectives, data selection strategies, and adaptive training schedules.

%% file: sections/Appendix.tex
\clearpage
\appendix
\section*{Appendix}
\section{Implementation Details}
\label{app:hyper-parameters}

This section summarizes the data construction procedure and the main hyperparameters used in the SFT warm-up and OPD stages. 
Unless otherwise specified, we use the same configurationsacross all experiments.

\paragraph{SFT data construction.}
To construct the SFT dataset, we randomly sample prompts from the OPD training set and use the teacher model to generate a response for each sampled prompt.
The decoding configuration used for teacher inference follows the base-model setting reported in Table~\ref{tab:inference-hyperparameters}.
The sampled prompts and their corresponding teacher-generated responses are then paired to form the SFT training data.

\paragraph{SFT warm-up.}
We consider both full-parameter fine-tuning and parameter-efficient fine-tuning with LoRA. 
The two settings use the same batch size and maximum sequence length, while different learning rates are adopted due to their different numbers of trainable parameters. 
The detailed hyperparameters are reported in Table~\ref{tab:sft-hyperparameters}.

\begin{table}[ht]
    \centering
    \small
    \begin{tabular}{lc}
        \toprule
        \textbf{Hyperparameter} & \textbf{Value} \\
        \midrule
        Batch Size                       & 16 \\
        Learning Rate (Full Fine-tuning) & \(5 \times 10^{-6}\) \\
        Learning Rate (LoRA)             & \(5 \times 10^{-5}\) \\
        \bottomrule
    \end{tabular}
    \caption{Hyperparameters used for SFT warm-up.}
    \label{tab:sft-hyperparameters}

\end{table}

\paragraph{OPD.}
After the SFT warm-up stage, we further optimize the student model using OPD. 
During training, the student generates one response for each prompt, while the teacher model provides token-level supervision for policy optimization. 
We use a maximum prompt length of 2,048 tokens and allow responses of up to 8,192 tokens. The main OPD hyperparameters are summarized in Table~\ref{tab:opd-hyperparameters}.

\begin{table}[ht]
    \centering
    \small
    \begin{tabular}{lc}
        \toprule
        \textbf{Hyperparameter} & \textbf{Value} \\
        \midrule
        Global batch Size                    & 128 \\
        Mini batch size               &128 \\
        Rollout \(n\)                 & 1 \\
        Maximum Prompt Length         & 2,048 \\
        Maximum Response Length       & 8,192 \\
        Temperature                   & 1.0 \\
        Top-\(p\)                     & 1.0 \\
        Learning Rate                 & \(1 \times 10^{-6}\) \\
        KL Coefficient                &0.0 \\
        \bottomrule
    \end{tabular}
    \caption{Hyperparameters used for OPD.}
    \label{tab:opd-hyperparameters}

\end{table}

\paragraph{Inference and evaluation.}
We use different decoding configurations for base and instruction-tuned models during evaluation.
The detailed inference hyperparameters are summarized in Table~\ref{tab:inference-hyperparameters}.

\begin{table}[ht]
    \centering
    \small
    \setlength{\tabcolsep}{5pt}
    \begin{tabular}{lccc}
        \toprule
        \textbf{Hyperparameter}
        & \textbf{Base}
        & \textbf{\shortstack{Instruct}}
        & \textbf{Thinking}\\
        \midrule
        do sample              & True    & True & True \\
        Max. Response Length & 16,384 & 24,576 & 32768 \\
        Temperature          & 0.7    & 0.6 & 0.6 \\
        Top-\(p\)             & 0.8    & 0.95 & 0.95 \\
        Top-\(k\)             & 20     & 20 & 20 \\
        \bottomrule
    \end{tabular}
    \caption{Inference hyperparameters used for model evaluation.}
    \label{tab:inference-hyperparameters}

\end{table}

\paragraph{Overlap ratio.}
For each student-generated context $c_t$, we obtain the top-$k$ token
sets $\mathcal{V}^{S}_t$ and $\mathcal{V}^{T}_t$ from the student and
teacher distributions, respectively. The overlap ratio is defined as
\begin{equation}
\mathrm{Overlap}
=
\frac{1}{|\mathcal{I}|}
\sum_{t\in\mathcal{I}}
\frac{
|\mathcal{V}^{S}_t \cap \mathcal{V}^{T}_t|
}{k},
\end{equation}
where $\mathcal{I}$ denotes all valid response positions. We use
$k=\textbf{32}$ and average the ratio over the training batch.
\label{app:overlap_ratio}

\section{Detailed Results for Warm-up Data}
\label{app:warm-up_data}

\subsection{Presence of CoT}
\label{app:cot_presence}
\begin{figure*}[t]
    \centering
    \includegraphics[width=\textwidth]{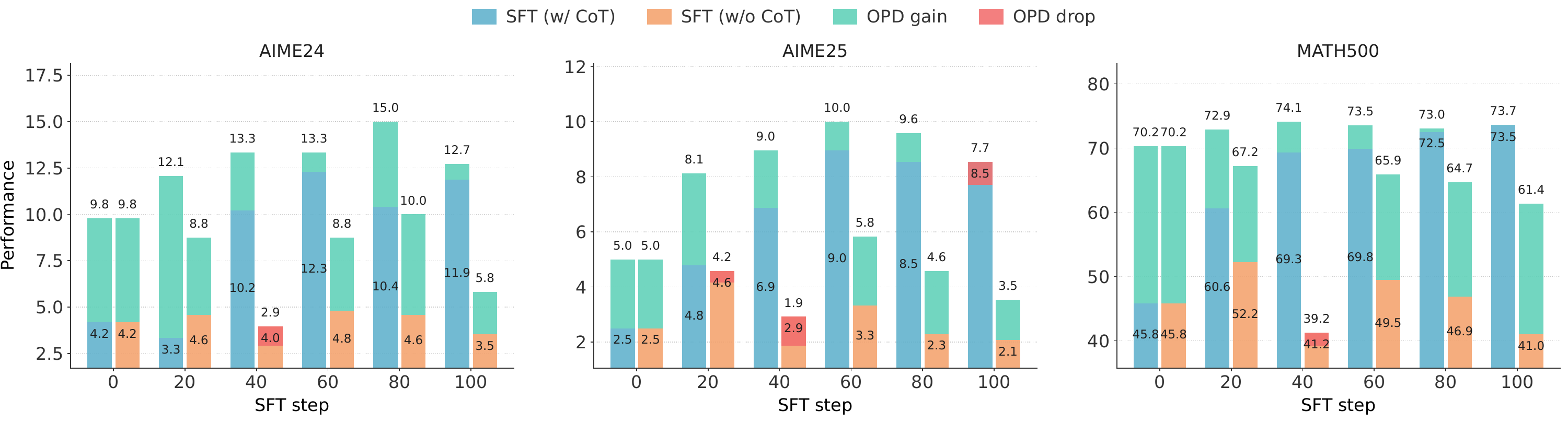}
\caption{
Detailed performance of SFT with and without CoT supervision across different SFT checkpoints. 
}
\label{fig:cot-detail}
\end{figure*}

\begin{figure*}[t]
    \centering
    \includegraphics[width=\textwidth]{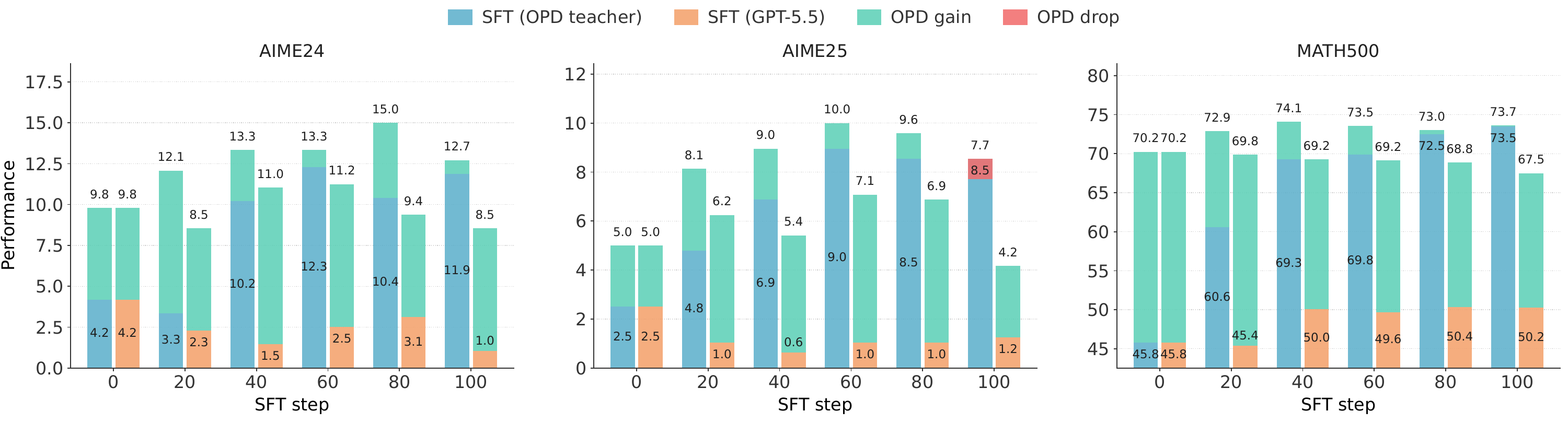}
\caption{
Performance comparison between SFT models trained with CoT responses generated by the OPD teacher and by an external teacher across different SFT checkpoints.
}
\label{fig:cot-source}
\end{figure*}

Figure~\ref{fig:cot-detail} compares SFT with and without CoT supervision across different SFT checkpoints.
Across the three benchmarks, SFT with CoT generally achieves substantially better performance than SFT without CoT.
This advantage is largely preserved after the subsequent OPD stage, especially at later SFT checkpoints.
For example, at SFT step 100, the CoT models achieve final OPD performances of 12.7, 8.5, and 73.7 on AIME24, AIME25, and MATH500, respectively, compared with 5.8, 3.5, and 61.4 for their non-CoT counterparts.
Moreover, OPD cannot consistently compensate for the absence of CoT supervision and occasionally even decreases performance on the AIME benchmarks.
These results demonstrate that CoT supervision is necessary for obtaining a strong and reliable initialization before OPD training.

\subsection{Source of CoT}
\label{app:cot_source}
Figure~\ref{fig:cot-source} compares SFT data generated by the OPD teacher with data generated by an external teacher.
Using CoT responses from the OPD teacher consistently yields a stronger SFT model across all three benchmarks.
This advantage is largely retained after OPD training, whereas OPD cannot fully compensate for the weaker initialization produced by the external teacher.
These results suggest that aligning the source of CoT supervision with the OPD teacher is important for effective policy distillation.

\begin{figure*}[ht]
    \centering
    \includegraphics[width=\textwidth]{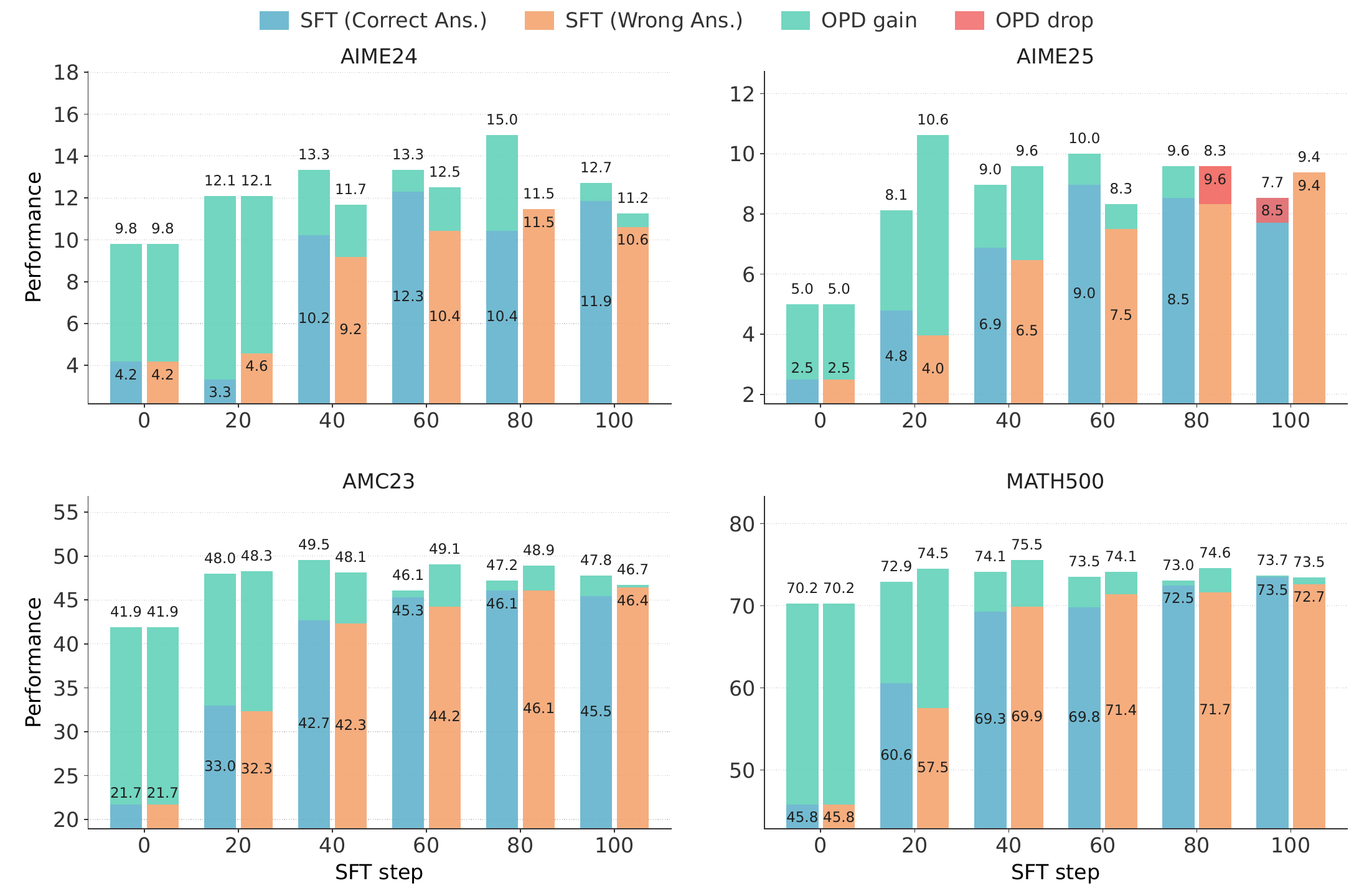}
    \caption{
    Performance comparison between SFT models trained with correct and incorrect CoT responses across different SFT checkpoints.
    }
    \label{fig:cot-correctness}
\end{figure*}

\begin{figure*}[ht]
    \centering
    \includegraphics[width=\textwidth]{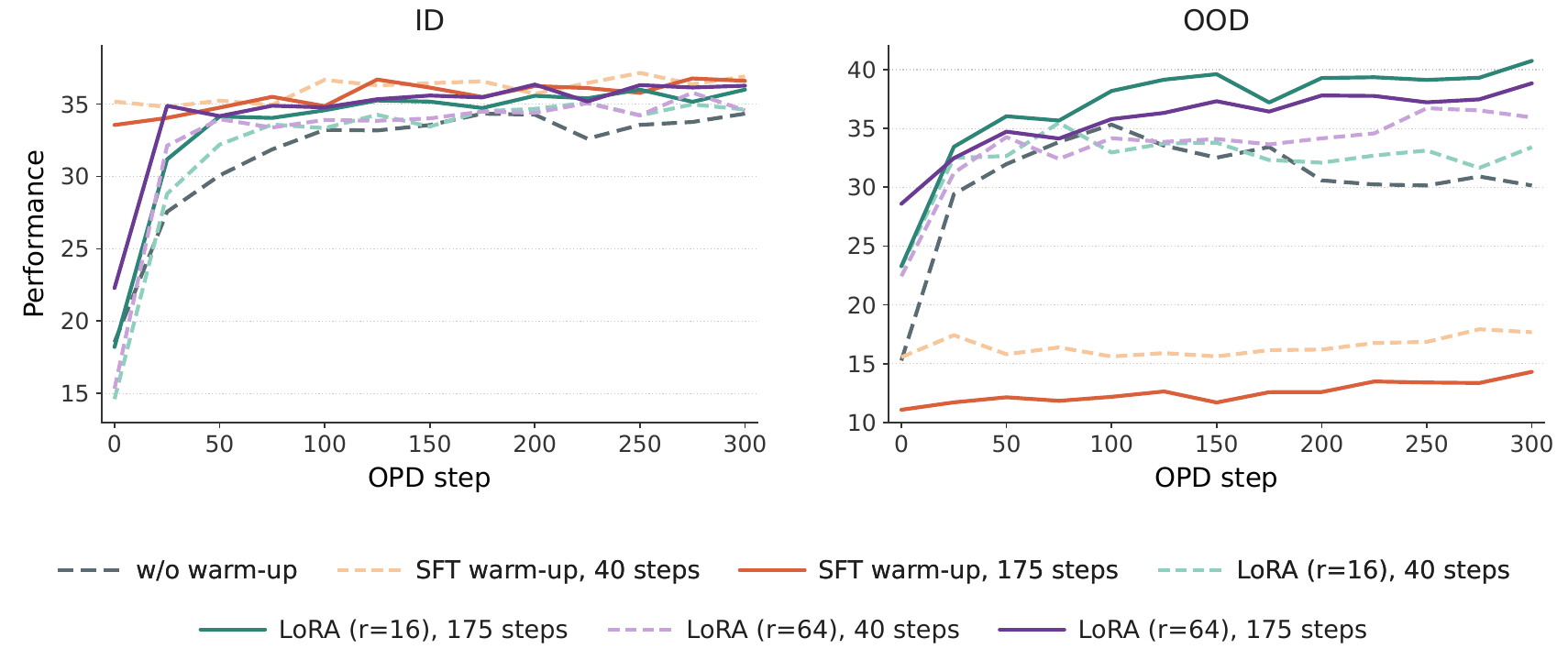}
    \caption{
    ID and OOD performance during OPD training under different SFT warm-up configurations.
    We compare full-parameter warm-up and LoRA warm-up with ranks \(r=16\) and \(r=64\), using either 40 or 175 SFT steps.
    Solid and dashed lines denote configurations with 175 and 40 warm-up steps, respectively.
    }
    \label{fig:lora-rank-warm-up}
\end{figure*}

\input{tables/same_size}

\input{tables/cot}

\subsection{Response Correctness}
\label{app:cot_correctness}

Figure~\ref{fig:cot-correctness} compares models trained with correct and incorrect CoT responses.
Across the four benchmarks, incorrect CoT supervision achieves performance comparable to correct CoT supervision after both SFT and OPD.
As illustrated in Table~\ref{tab:paired_similar_reasoning}, an incorrect response can preserve nearly the same thinking pattern as a correct response, including the same problem decomposition, intermediate computations, and self-checking structure.
The incorrect trajectory differs only in a local arithmetic error, which is subsequently propagated through the remaining steps.
These results suggest that preserving an effective reasoning pattern may be more important than ensuring that every SFT trajectory is fully correct.

\section{Detailed Results for Warm-up Training}
\label{app:lora-rank-warm-up}
\paragraph{Effect of LoRA rank and warm-up steps.}
Figure~\ref{fig:lora-rank-warm-up} studies the effects of the LoRA rank and the number of SFT warm-up steps.
Increasing the LoRA rank provides only limited improvements on the in-domain benchmarks.
On the out-of-domain benchmarks, a smaller LoRA rank can even achieve better performance, indicating that a larger rank does not necessarily improve generalization.
In contrast, increasing the number of SFT warm-up steps leads to a clear and consistent improvement, particularly on the out-of-domain benchmarks.
These results suggest that the warm-up duration is more important than the LoRA rank for obtaining a strong initialization for OPD.

\input{tables/case_study2}

%% file: tables/same_size.tex
\begin{table*}[t]
\centering
\setlength{\tabcolsep}{3.0pt}
\renewcommand{\arraystretch}{1.12}
\begin{tabular}{@{}lccccc@{\hspace{6pt}}cccccc@{}}
\toprule
\textbf{Method}
& \multicolumn{5}{c}{\textbf{ID}}
& \multicolumn{6}{c}{\textbf{OOD}} \\
\cmidrule(lr){2-6}
\cmidrule(lr){7-12}
& \makecell{\textbf{AIME}\\\textbf{24}}
& \makecell{\textbf{AIME}\\\textbf{25}}
& \makecell{\textbf{AMC}\\\textbf{23}}
& \makecell{\textbf{MATH}\\\textbf{500}}
& \textbf{Avg.}
& \textbf{IFEval}
& \textbf{GPQA}
& \textbf{Phys.}
& \textbf{Chem.}
& \textbf{Hist.}
& \textbf{Avg.} \\
\midrule

Teacher
& 52.50
& 39.58
& 91.25
& 91.60
& 68.73
& 29.94
& 31.31
& 50.73
& 47.00
& 14.44
& 34.68 \\
\midrule

Student
& 25.83
& 22.08
& 66.56
& 83.90
& 49.59
& 29.02
& 13.64
& 40.95
& 37.72
& 13.39
& 26.94 \\
\midrule

OPD
& 43.33
& 33.75
& 84.69
& \textbf{89.10}
& 62.72
& \textbf{29.76}
& \textbf{26.26}
& \textbf{48.34}
& 43.99
& 13.39
& \textbf{32.35} \\
\rowcolor{blue!5}
Simple-OPD
& \textbf{45.42}
& \textbf{36.67}
& \textbf{86.56}
& 88.70
& \textbf{64.34}
& 29.02
& 25.25
& 44.42
& \textbf{45.49}
& \textbf{15.49}
& 31.93 \\
\bottomrule
\end{tabular}

\caption{
Detailed benchmark-level results with a same-size student-trained teacher.
The Simple-OPD row is shaded, and the better result between OPD and Simple-OPD is shown in bold.
}
\label{tab:same_size}
\end{table*}

%% file: tables/cot.tex
\begin{table*}[t]
\centering
\footnotesize
\setlength{\tabcolsep}{6pt}
\renewcommand{\arraystretch}{1.15}
\begin{tabular}{@{}p{0.98\linewidth}@{}}
\toprule
\textbf{Question.}\;
How many positive integers less than $4000$ are not divisible by
$2$, $3$, $5$, or $7$?\quad
\textbf{Gold:} $913$.
\smallskip
\\
\midrule
\textbf{Correct response.}\quad Predicts $\boxed{913}$\;\checkmark
\\
\midrule
\emph{``To find the number of integers not divisible by $2$, $3$,
$5$, or $7$, it might be easier to use the principle of
inclusion--exclusion.''}\\[2pt]
\emph{\textbf{Step 1: Find the total number of positive integers less
than $4000$.}}
\emph{The range is $1,\ldots,3999$, so the total is $3999$.}\\[2pt]
\emph{\textbf{Step 2: Find the number of integers less than $4000$
that are divisible by $2$, $3$, $5$, or $7$.}}
\emph{Define $A,B,C,D$ as the sets of multiples of $2,3,5,7$ and
apply inclusion--exclusion.}\\[2pt]
\emph{Single-set counts:}
\emph{$|A|=1999$, $|B|=1333$, $|C|=799$, and $|D|=571$;
their sum is $4702$.}\\[2pt]
\emph{Pairwise counts:}
\emph{$|A\cap B|=666$, $|A\cap C|=399$, $|A\cap D|=285$,
$|B\cap C|=266$, $|B\cap D|=190$, and $|C\cap D|=114$;
their sum is $\mathbf{1920}$.}\\[2pt]
\emph{Triple counts:}
\emph{$133,95,57,38$, whose sum is $323$; the four-way
intersection has size $19$.}\\[2pt]
\emph{Hence
$|A\cup B\cup C\cup D|=4702-1920+323-19=3086$.}\\[2pt]
\emph{\textbf{Step 3: Find the number of integers less than $4000$
that are not divisible by $2$, $3$, $5$, or $7$.}}
\emph{$3999-3086=\mathbf{913}$.}\\[2pt]
\emph{``Wait, let's verify the calculations to ensure accuracy.''}
\emph{The response repeats the four inclusion--exclusion totals and
checks $3999-3086=913$.}\\[2pt]
\emph{Answer: $\boxed{913}$.}
\\
\midrule
\textbf{Incorrect response.}\quad
Predicts $\textcolor{red}{\boxed{1013}}$\;$\textcolor{red}{\times}$
\\
\midrule
\emph{``To find the number of positive integers less than $4000$
that are not divisible by $2$, $3$, $5$, or $7$, it might be easier
to use the principle of inclusion--exclusion.''}\\[2pt]
\emph{\textbf{Step 1: Find the total number of positive integers less
than $4000$.}}
\emph{The range is again $1,\ldots,3999$, giving $3999$.}\\[2pt]
\emph{\textbf{Step 2: Find the number of positive integers less than
$4000$ that are divisible by $2$, $3$, $5$, or $7$.}}
\emph{The response defines the same four sets and writes the same
inclusion--exclusion formula.}\\[2pt]
\emph{It obtains the same single-set counts
$1999,1333,799,571$ (sum $4702$), the same pairwise counts
$666,399,285,266,190,114$, the same triple counts
$133,95,57,38$ (sum $323$), and the same four-way count $19$.}\\[2pt]
\emph{It then computes}
\emph{$4702-(666+399+285+266+190+114)
=4702-\textcolor{red}{\mathbf{2020}}=2682$.}\\[1pt]
\textcolor{purple}{\textbf{[Arithmetic slip:}}
\textcolor{purple}{the six displayed pairwise counts sum to
$1920$, not $2020$.\textbf{]}}\\[2pt]
\emph{Following the same remaining steps, it obtains
$2682+323-19=2986$ and then
$3999-2986=\textcolor{red}{\mathbf{1013}}$.}\\[2pt]
\emph{``Wait, let's double-check the calculations to ensure
accuracy.''}
\emph{The check repeats the same incorrect subtotal $2020$ rather
than independently re-adding its six terms. The response even notes
that an alternative cross-check would help, but proceeds without one.}\\[2pt]
\emph{Answer: $\textcolor{red}{\boxed{1013}}$.}
\\
\bottomrule
\end{tabular}
\caption{
Correct and incorrect responses with similar reasoning patterns.
Both responses follow the same inclusion--exclusion procedure and self-checking structure, while the incorrect response makes a local error when summing the pairwise intersections and propagates it to the final answer.
}
\label{tab:paired_similar_reasoning}
\end{table*}

%% file: tables/case_study2.tex
\begin{table*}[t]
\centering
\footnotesize
\setlength{\tabcolsep}{6pt}
\renewcommand{\arraystretch}{1.15}
\begin{tabular}{@{}p{0.98\linewidth}@{}}
\toprule
\textbf{Prompt.}\;
What is the sum of all of the multiples of $7$ between $100$ and $200$?\quad
\textbf{Gold:} $2107$.
\smallskip
\\
\midrule
\textbf{Teacher response.} Predicts $\boxed{2107}$\;\checkmark
\\
\midrule
\emph{Alright, I have this problem to solve. [\ldots]}\\[2pt]
\emph{Step 1: Identify the smallest multiple of $7$ greater than $100$.}
\emph{$100/7\approx 14.29$, round up to $15$, so
$7\times 15=105$.}\\[2pt]
\tea{\textbf{Verification:}}
\emph{Is $105>100$? Yes. Is $105$ a multiple of $7$? Yes.}\\[2pt]
\emph{Step 2: Identify the largest multiple of $7$ less than $200$.}
\emph{$200/7\approx 28.57$, round down to $28$, so
$7\times 28=196$.}\\[2pt]
\tea{\textbf{Verification:}}
\emph{Is $196<200$? Yes. Is $196$ a multiple of $7$? Yes.}\\[2pt]
\emph{Step 3: List all multiples of $7$ from $105$ to $196$.}
\emph{Arithmetic sequence with $a_{1}{=}105$, $d{=}7$, $a_{n}{=}196$.}\\[2pt]
\emph{Step 4: Determine the number of terms.}
\emph{$196=105+(n-1)\cdot 7\Rightarrow n=14$.}\\[2pt]
\emph{Step 5: Calculate the sum.}
\emph{$S_{14}=\tfrac{14}{2}\times(105+196)=7\times 301=\mathbf{2107}$.}\\[2pt]
\tea{\textbf{Verification}: Let's verify the calculations to ensure accuracy.}
\emph{(a) List all $14$ terms $105,112,119,\ldots,189,196$ and
count: $14$ terms, correct. (b) Pair them:
$105{+}196=301$, $112{+}189=301$, $119{+}182=301$, $\ldots$; each
pair sums to $301$; $7$ pairs. Total
$=7\times 301=\mathbf{2107}$. Correct.}\\[2pt]
\emph{Final Answer:} $\boxed{2107}$.
\\
\midrule
\textbf{Warm-up response.}\quad Predicts $\boxed{2107}$\;\checkmark
\\
\midrule
\emph{Alright, I have this problem to solve. Alright, let's break
this down. [\ldots Understanding the Problem \ldots]}\\[2pt]
\emph{Step 1: Identify the First Multiple of $7$ Greater Than $100$.}
\emph{$100/7\approx 14.29$, round up to $15$, so
$7\times 15=105$.}\\[2pt]
\emph{Step 2: Identify the Last Multiple of $7$ Less Than $200$.}
\emph{$200/7\approx 28.57$, round down to $28$, so
$7\times 28=196$.}\\[2pt]
\emph{Step 3: List All Multiples of $7$ Between $100$ and $200$.}
\emph{$105,112,\ldots,196$. ``\tea{Wait, let's verify the range}''
--- re-lists to confirm.}\\[2pt]
\emph{Step 4: Calculate the Number of Terms in the Sequence.}
\emph{$196=105+(n-1)\cdot 7\Rightarrow n=14$.}\\[2pt]
\emph{Step 5: Calculate the Sum of the Arithmetic Sequence.}
\emph{$S_{14}=\tfrac{14}{2}\times(105+196)=7\times 301=\mathbf{2107}$.}\\[2pt]
\tea{\textbf{Verification.}}
\emph{``Let's verify the sum by adding all the multiples of $7$
from $105$ to $196$.'' Manual running sum:
$105{+}112=217$;\;$217{+}119=336$;\;$336{+}126=462$;\;
$462{+}133=595$;\;$\ldots$;\;$1722{+}189=1911$;\;
$1911{+}196=\mathbf{2107}$. \emph{``The sum is indeed
$2107$.''}}\\[2pt]
\emph{Conclusion: $\boxed{2107}$.}
\\
\midrule
\textbf{No-warm-up response.}\quad Predicts $\err{\boxed{1407}}$\;$\err{\times}$
\\
\midrule
\emph{To solve the problem, we follow these steps: [numbered
list \ldots]}\\[2pt]
\emph{\#\#\# Step 1: Identify the smallest multiple of $7\geq 100$.}
\emph{$100/7\approx 14.29$, round up to $15$, so
$7\times 15=105$.}\\[2pt]
\emph{\#\#\# Step 2: Identify the largest multiple of $7\leq 200$.}
\emph{$200/7\approx 28.57$, round down to $28$, so
$7\times 28=196$.}\\[2pt]
\emph{\#\#\# Step 3: Determine the number of terms in this sequence.}
\emph{$196=105+(n-1)\cdot 7\Rightarrow n=14$.}\\[2pt]
\emph{\#\#\# Step 4: Calculate the sum of the arithmetic sequence.}
\emph{$S_{14}=\tfrac{14}{2}\times(105+196)=7\times
\err{201}=\err{1407}$.
\quad\textcolor{purple}{[Arithmetic slip: $105+196=301$, not $\err{201}$.]}}\\[2pt]
\err{\textbf{[No Verification step.]}}
\emph{The response ends here---no re-check of $a_{1}+a_{n}$, no
manual re-sum of the $14$ terms, no sanity check that the sum
of $14$ three-digit numbers must exceed $14\times 105=1470$.}\\[2pt]
\emph{Final Answer: $\err{\boxed{1407}}$.}
\\
\bottomrule
\end{tabular}
\caption{
Response skeletons for MATH500 problem at
OPD step~\(300\).
The teacher and warm-up responses independently verify the
arithmetic-sequence sum, whereas the no-warm-up response makes the
error \(105+196=201\) and terminates without checking it.
Step headings and quoted phrases are verbatim, while the accompanying
descriptions are condensed.
}
\label{tab:case_pattern}
\end{table*}